# HAR-Net:Fusing Deep Representation and Hand-crafted Features for Human Activity Recognition

Mingtao Dong[1], Jindong Han[2]

[1] The Second High School attached to Beijing Normal University, Beijing, China
`mingtaod@andrew.cmu.edu`
[2] Beijing University of Posts and Telecommunications, Beijing, China
`hanjindong@bupt.edu.cn`

**Abstract.** Wearable computing and context awareness are the focuses of study in the field of artificial intelligence recently. One of the most appealing as well as challenging applications is the Human Activity Recognition (HAR) utilizing smart phones. Conventional HAR based on Support Vector Machine relies on subjective manually extracted features. This approach is time and energy consuming as well as immature in prediction due to the partial view toward which features to be extracted by human. With the rise of deep learning, artificial intelligence has been making progress toward being a mature technology. This paper proposes a new approach based on deep learning and traditional feature engineering called HAR-Net to address the issue related to HAR. The study used the data collected by gyroscopes and acceleration sensors in android smart phones. The raw sensor data was put into the HAR-Net proposed. The HAR-Net fusing the hand-crafted features and high-level features extracted from convolutional network to make prediction. The performance of the proposed method was proved to be 0.9% higher than the original MC-SVM approach. The experimental results on the UCI dataset demonstrate that fusing the two kinds of features can make up for the shortage of traditional feature engineering and deep learning techniques.

## 1 Introduction

With the advent of big data era, it has been an urge for humans to find valuable information. Among all types of applications addressing this issue, Human Activity Recognition (HAR), which is applied frequently to the fields of human-computer interaction, identification technology, and medical security, has been the one that is commonly utilized. Currently, studies related to HAR is based on several approaches: video-based HAR and wearable sensor-based HAR. The former approach of HAR is based on the analysis and recognition videos captured by camera from moving human bodies. This approach is mainly employed by the field of security monitoring, being especially useful when monitoring anomalous behavior of the elderly and children. Despite the multiple applications and benefits of video-based HAR, there are challenges that may hinder the development of the technology. Due to the demand for storage resources and the limitations related to the deployment of cameras, the monitoring of users' whole-day activity state is impossible. This restriction causes the considerable difficulty of popularization for the video-based approach.

Wearable sensor-based HAR, as an area of intensive interest, on the other hand, can be achieved using wearable acceleration sensor, gyroscope, and heart rate sensor. The potential of this approach is enabled by the portability and low cost of wearable sensors. As an intelligent aided technology, sensor-based HAR has been employed into commercialized devices as fitness tracker and fall detector. More socially valuable applications, like providing dementia patients with Memory Prosthesis and proposing suggestions for daily exercise which require a comprehensive understanding about past and present human activity, are expected to be generalized. Whereas the traditional approach of wearable-based HAR can be challenging to generalize as the result of the requirement for specially designed sensors. In the 1950s, there were researchers related to sensor-based HAR whose progress was retarded due to the immaturity of recognition algorithm and unaffordable huge sized sensors. Decades of following advancement in microelectronics and computer motivated the invention of MEMS that permitted the micromotion and integration of sensors. Sensors now have attributes of high computational power, low cost, and tiny size. Meanwhile, the popularization of smart phone enables people to enjoy an intelligent platform that incorporates amusement, communication, and work,it counteracts the toughness faced by wearable-based HAR. With the rise of artificial intelligence and wearable computing, plenty of sensors have been integrated into smart phones. HAR based on smart phone built-in sensors has thus become research hotspots. The portability and the ease to develop software on smart phone makes real-time supervision of human body activity and generalization of smart phone-based HAR applications feasible. Real-time supervision quantifies human activity, raises humans' self-awareness of health conditions, and offers reasonable fitness tips, all of which make it a promising market with high value.

The rest of this paper is organized as follows: Section 2 summarizes the related work. The details of our model are illustrated in Section 3. Our model will be evaluated in Section 4. We draw conclusions in section 5.

## 2    Related Work

With the hardware development, sensor-based HAR has been commercially applied to military, medical system, and security. Massachusetts Institute of Technology (MIT) and University of Cambridge have developed LiveNet system intended to realize HAR in soldiers' training for military security [1]. Information collected from sensors is a time series. Different effective approaches to process time series have been proposed to act as the foundation of HAR. The majority of focus in HAR study covers daily life, sport, socialization, and health. Layering classification is achieved by precepting group behavior using multiple sensors, while children's socializing behaviors are successfully identified through physiological sensors [2] [3]. Through using acceleration sensor, researchers have achieved Parkinson patients' recognition and detection of Down's Syndrome patients [4] [5]. Researchers in Stanford have utilized multi-model sensors system to detect whether the elder is falling down [6]. Wearable sensors are also applied to the recognition of newborns' disorder related to cerebral stroke [7]. Sensor-based HAR can be optimistic prospect when used to support vulnerable groups.

Researchers have done study using an efficient Support Vector Machine algorithm to compensate for computational loss [8]. In recent years, deep learning is rising because of the big data. While conventional machine learning approach required researchers to manually extract features, deep learning can accomplish end to end learning, using original data as input without manual feature extraction. Deep neural network can provide powerful feature representation by complex non-linear transformation. The mostly used networks are deep convolutional neural networks and LSTM for video-based HAR. Researchers have shown the high competence of LSTM to classify activity appear in video by analyzing Sports-1 M and UCF-101 data set [9]. Utilizing three wearable sensor data set to assess the performance on deep learning framework, researchers are able to determine the effectiveness of deep learning on extracting features [10]. Additionally, conventional machine learning approach has been compared to sensor-based deep learning approach for human activity recognition [11].

When we use image and signals as input, convolutional neural networks perform better than multilayer neural networks. A study has been done by researchers with two dimensional images that are synthesized with 9 types of sensor data, processed by Discrete Fourier Transformation (DFT), and recognized with convolutional neural network [12]. Researchers have also adopted several convolutional neural networks trained in parallel followed by max pooling, concatenated the activations, and fed them into the fully connected layer. Convolutional neural networks can be advantageous when applied to HAR because of the local dependency and scale invariance [13].

## 3    Methodology

For the wearable sensor-based human activity recognition task,we proposed a network architecture called HAR-Net. The network architecture is illustrated in Figure 1. The input of the network will have a dimension of (1, 128, 9) with 9 channels,then will be processed by the separative convolutional layer. This means that the 9 channels of the input will be convoluted respectively by the same filters. For each channel of the input, it will be processed in parallel by filters with size (1, 1), (5, 1), (9, 1), (13, 1), (17, 1), (21, 1), (25, 1), (29, 1), and (33, 1)respectively, each followed by a max pooling layer with size (11, 1). The pattern of convolutional layer followed by a max pooling layer described above will be repeated for four times consecutively. Hand-crafted features,such as time and frequency domain statistical features, will be concatenated with the flattened output from the convolutional layers. Then fed into the first fully connected layer with 2048 neurons with relu activation function.The output of the first fully connected layer will be input into the second layer with 64 neurons and tanh activation function. Softmax is applied to the output of the second fully connected layer to classify different kinds of activities. Adam optimizer with default parameters [14] was applied to the HAR-Net.

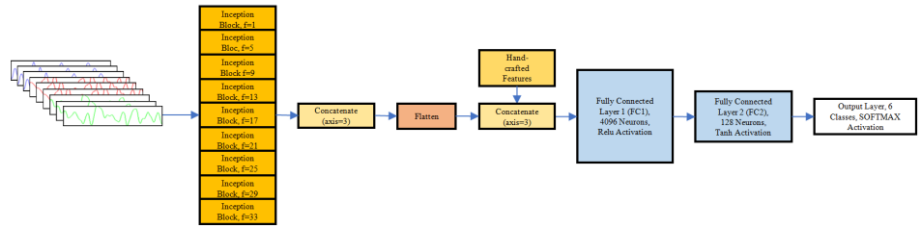

**Figure 1:** HAR-NetArchitecture.

The structure of the Inception Block is shown in Figure 2. Figure 3 shows the illustration of Separable Convolution. Separative convolution is chosen because of its advantage of preventing the interference between different channels of input.

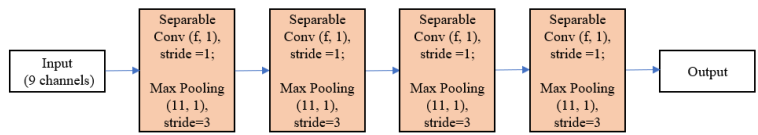

**Figure 2:** Inception Block with separable convolution.

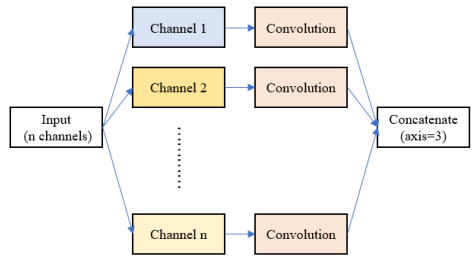

**Figure 3:** Separable Convolution

## 4 Experimental Results

### 4.1 HAR Dataset Description

The data adopted in this paper is a public domain dataset from UCI [15]. The subject of the experiment used to collect the data set has chosen to be a group of 30 people with age between 19 to 48 years old. Each person was asked to carry out the six specified Activities: walking downstairs, walking upstairs, sitting, laying, walking, and standing. The data in the process was collected using build-in accelerometer and gyroscope in smart phone at a constant sampling frequency of 50 Hz.

### 4.2 Results on UCI dataset

In order to evaluate the performance of the method we proposed, we compare out method with the following two baselines:

MC-SVM [15]:it was proposed by researchers from Universit`adegliStudi di Genova. This approach adopted the One-Versus-All (OVA) approach in multiclass SVM and used 561 manually extracted features as input.

The HAR-Net without Hand-crafted features:it was a modified version of the proposed HAR-Net model. The hand-crafted features were removed. And we fine-tune the previous model make it more suitable for this task.

| Approach | Prediction Accuracy |
|---|---|
| MC-SVM | 96.0% |
| HAR-Net with Hand-crafted Features | 96.9% |
| HAR-Net withoutHand-crafted Features | 93.5% |

**Table 1:**Comparison among different methods on UCI dataset

Table 1 shows the results of our model and other baselines on UCI dataset.According to the study proposing the MC-SVM approach, among the 2,947 test set samples, the overall accuracy of prediction can reach 96.0%. Despite the high accuracy reached, there are potential improvements for the approach. According to the HAR-Net proposed in this paper, convolutional neural network can automatically select the most effective features for higher classification accuracy through utilizing loss function and back propagation. Through applying different filters to the signal, the deep neural network is able to detect similar features in signal waveform and generate distinctive feature maps. When the network gets deeper, non-linearity of the neural network increases. More abstract features areextracted by the neural network adaptively through the automatic learning process. The advantage of the features extracted through deep learning is the decreaseof the amount ofhuman's subjective judgement about which features to use. The HAR-Net combined the Hand-crafted features with the learnt features by the deep neural network to provide more comprehensive perspectives when making predictions. Regarding the prediction accuracy for the HAR-Net, the overall prediction accuracy reached 96.9% and outperformed the MC-SVM model by 0.9%,demonstrating the effectiveness of the HAR-Net.

Meanwhile, we compare the HAR-Net withthe network that didn't use the hand-crafted features. The network without manually extracted statistics features may fail to capture the extra effective representation and the performance is worse than the HAR-Net. This can further prove the importance of adding the auxiliary feature engineering.

| Actual Class | Predicted Class | | | | | |
|---|---|---|---|---|---|---|
| | Walking | Walking Upstairs | Walking Downstairs | Sitting | Standing | Laying |
| Walking | 490 | 0 | 6 | 0 | 0 | 0 |
| Walking Upstairs | 14 | 454 | 3 | 0 | 0 | 0 |
| Walking Downstairs | 4 | 9 | 407 | 0 | 0 | 0 |
| Sitting | 0 | 3 | 0 | 445 | 43 | 0 |
| Standing | 0 | 0 | 0 | 10 | 522 | 0 |
| Laying | 0 | 0 | 0 | 0 | 0 | 537 |

**Table 2:** Confusion matrix of HAR-Net

The confusion matrix of the HAR-Netis shown above in Table 2. Despite the overall high accuracy, confusions happened when the model was used to predict activities. From the confusion matrix above, a pattern of mild confusion between walking, walking upstairs, and walking downstairs appears. A more serious confusion between standing and sitting is discovered due to thelarge number of sitting that were misidentified as standing. The confusion between those activities were caused by the similarity in the signal' frequency and amplitude.

### 4.3 Separable Convolution versusConventional Convolution

A set of comparative experiments were done to illustrate the choice of Separable Convolution instead of default convolutional layer. The Inception Block for default convolution approach is shown below in Figure 4.

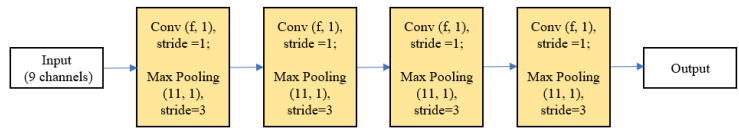

**Figure 4:** The HAR-Net without separable convolution

For conventional convolutional layer, the output channel size is equal to the number of filters applied. For each filter applied, the convolutions of the all the channels are added together to generate a single channel output. The prediction accuracy comparison is shown in Table 3.

| Approach | Prediction Accuracy |
|---|---|
| HAR-Net based on Conventional Convolution | 95.2% |
| HAR-Net based on Separable Convolution | 96.9% |

**Table 3:** Comparison among the separable conv-based HAR-Net and conventional conv-based HAR-Net

As shown in Table 3, the HAR-Net based on separable convolution outperformed the convention convolution-based model by 1.7%. This indicate the advantage of separable convolution over conventional convolution approach in processing 1-dimensional time series signal. The approach of separable convolution reduces the interference between data from different channels (axis). It retains the features extracted from separate channels apart to provide more valid features available for automatic learning in the deep neural network. The confusion matrix of model with conventional convolution is shown in Table 4 below.

| Actual Class | | Predicted Class | | | | | |
|---|---|---|---|---|---|---|---|
| | | Walking | Walking Upstairs | Walking Downstairs | Sitting | Standing | Laying |
| | Walking | 481 | 0 | 15 | 0 | 0 | 0 |
| | Walking Upstairs | 1 | 454 | 16 | 0 | 0 | 0 |
| | Walking Downstairs | 3 | 5 | 412 | 0 | 0 | 0 |
| | Sitting | 0 | 2 | 0 | 429 | 57 | 3 |
| | Standing | 1 | 0 | 0 | 37 | 494 | 0 |
| | Laying | 0 | 1 | 0 | 0 | 0 | 536 |

**Table 4:** Confusion matrix ofthe model using conventional convolution

From Table 4 above, serious confusions appeared when distinguishing sitting and standing. The model leveraging the separable convolution performed better when distinguishing sitting and standing as well as walking and walking downstairs.

## 5 Conclusions

In the study, a new approach utilizing Inception Block for convolutional neural network with hand-crafted features was proposed to address Human Activity Recognition. The HAR-Net for convolutional neural network can be an appealing option for realizing HAR in health care industry to take care of the vulnerable groups. Due to the advantages of convolutional neural network's local dependency and scale invariance [13], it can be a potential application in the field of HAR. The HAR-Net outcompetes the traditional MC-SVM approach because of the higher prediction accuracy resulted from the extra self-adaptive features extracted by deep learning in the model.

The study confirmedthat the combination of deep learning techniques and traditional feature engineering can outcompete MC-SVM approach using the same public domain data set to the prediction accuracy. The result suggests an implication about the smartphone's effect on recognizing human activity. Future work will include testing with the Residual Network (Resnet) and Recurrent Network to realize HAR utilizing more datasets and reduce the degree of confusion between distinctive activities especially sitting and standing.

## References

1. Sung, M.: LiveNet: Health and Lifestyle Networking Through Distributed Mobile Devices. pp. 15-17. In: Proceedings WAMES, Boston (2004).
2. Hernandez, J., Riobo, I., Rozga, A., Abowd, G.D., Picard, R.W.: Using Electrodermal Activity to Recognize Ease of Engagement in Children during Social Interactions. In: ACM International Joint Conference on Pervasive & Ubiquitous Computing, vol. 48, pp. 307-317. ACM, Geneva (2014).
3. Mikkel Baun Kjærgaard, Wirz, M., & Roggen, D.: Detecting Pedestrian Flocks by Fusion of Multi-Modal Sensors in Mobile Phones. In: ACM Conference on Ubiquitous Computing, pp. 240-249. ACM, Geneva (2012).
4. Chengqing, Z.: Statistical Natural Language Processing. Tsinghua University Press, Beijing (2008).
5. Xinqing S.: A Brief Treatise on Computational Electromagnetics. Press of University of Science and Technology of China, Beijing (2004).
6. Murata, S., Suzuki, M., &Fujinami, K.: A Wearable Projector-based Gait Assistance System and its Application for Elderly People. In: ACM International Joint Conference on Pervasive & Ubiquitous Computing, pp. 143-152. ACM, Geneva (2013).
7. Fan, M., Gravem, D., Dan, M. C., Patterson, D. J.: Augmenting Gesture Recognition with Erlang-Cox Models to Identify Neurological Disorders in Premature Babies. In: International Conference on Ubiquitous Computing, pp. 411-420. ACM. Geneva(2012).
8. Anguita, D., Ghio, A., Oneto, L., Parra, X., Reyes-Ortiz, J. L.: Energy Efficient Smartphone-Based Activity Recognition using Fixed-Point Arithmetic. Journal of Universal Computer Science 19(9), 1295-1314(2013).
9. Ng, Y. H., Hausknecht, M., Vijayanarasimhan, S., Vinyals, O., Monga, R., &Toderici, G.: Beyond Short Snippets: Deep Networks for Video Classification. Computer Vision and Pattern Recognition 16(4), 4694-4702 (2015).
10. Plötz, T.,Hammerla, N. Y., Olivier, P.: Feature Learning for Activity Recognition in Ubiquitous Computing. In: Proceedings of the 22nd International Joint Conference on Artificial Intelligence (IJCAI), pp. 1729–1734. IJCAI, Barcelona,16–22 (2011).
11. Jiang, W., Yin, Z.: Human Activity Recognition using Wearable Sensors by Deep Convolutional Neural Networks. In: Proceedings of the 23rd ACM International Conference on Multimedia. ACM, Geneva, 1307-1310 (2015).
12. Cortes, C., Vapnik, V.: Support-vector networks. Machine learning, 20(3), 273–297 (1995).
13. Zeng, M., Le, T. N., Yu, B., Mengshoel, O. J., Zhu, J., Wu, P.: Convolutional Neural Networks for Human Activity Recognition using Mobile Sensors. In: International Conference on Mobile Computing, Applications and Services, pp.197-205. IEEE, New York (2015).
14. D. P. Kingma and J. Ba, "Adam: A method for stochastic optimization," CoRR, vol. abs/1412.6980, 2014. [Online]. Available: http://arxiv.org/abs/1412.6980
15. Anguita, D., Ghio, A., Oneto, L., Parra, X., Reyes-Ortiz, J. L.: A Public Domain Dataset for Human Activity Recognition Using Smartphones. In: 21th European Symposium on Artificial Neural Networks, Computational Intelligence and Machine Learning, pp.24-26. ESANN, Bruges (2013).